\documentclass[letterpaper]{article}
\usepackage{aaai}
\usepackage{times}
\usepackage{amsmath}
\usepackage{amsfonts}
\usepackage{helvet}
\usepackage{courier}
\usepackage{subfig}
\usepackage{adjustbox}
\usepackage{float}
\usepackage{url}
\usepackage{multirow}
\usepackage{subfig}
\usepackage{graphicx}
\usepackage{booktabs}
\usepackage{xcolor}
\usepackage{nameref}

\begin{document}
%
\title{Interpreting Deep Neural Networks for Medical Imaging using Concept Graphs}
\author{ Avinash Kori\\
  koriavinash1@gmail.com\\ 
  Department of Engineering Design\\
  Indian Institute of Technology, Madras\\
\And
Parth Natekar\\
patnat26@gmail.com\\
Department of Engineering Design\\
Indian Institute of Technology, Madras\\
\AND
Ganapathy Krishnamurthi \\
gankrish@iitm.ac.in\\
Department of Engineering Design\\
Indian Institute of Technology, Madras\\
\And
Balaji Srinivasan \\
sbalaji@iitm.ac.in\\
Department of Mechanical Engineering\\
Indian Institute of Technology, Madras\\
}
\maketitle
\begin{abstract}
\begin{quote}
The black-box nature of deep learning models prevents them from being completely trusted in domains like biomedicine. Most explainability techniques do not capture the concept-based reasoning that human beings follow. In this work, we attempt to understand the behavior of trained models that perform image processing tasks in the medical domain by building a graphical representation of the concepts they learn. Extracting such a graphical representation of the model's behavior on an abstract, higher conceptual level would help us to unravel the steps taken by the model for predictions. We show the application of our proposed implementation on two biomedical problems - brain tumor segmentation and fundus image classification. We provide an alternative graphical representation of the model by formulating a \textit{concept level graph} as discussed above, and find active inference trails in the model. We work with radiologists and ophthalmologists to understand the obtained inference trails from a medical perspective and show that medically relevant concept trails are obtained which highlight the hierarchy of the decision-making process followed by the model. 
Our framework is available at \url{https://github.com/koriavinash1/BioExp}.
\end{quote}
\end{abstract}
\section{Introduction}\label{sec:intro}

Deep learning models are black boxes and as they are integrated into medical diagnosis, it becomes necessary to give a clear explanation of the concepts learnt by the model in a form understandable to medical professionals \cite{holzinger2017we}. Clinicians also prefer upfront information about the global properties of a model, such as its known strengths and limitations \cite{cai2019hello}. 

For this, semantic concepts internal to the model and their relationships need to be identified and represented in a human-understandable form. Previous interpretability techniques are example based or attention based \cite{molnar2020interpretable}, such as attribution, saliency, or feature visualization, and do not reflect the 'concept-based thinking' that human-reasoning shows \cite{armstrong1983some}, neither do they allow us to uncover the model's understanding of the relationship between such concepts. In the related work section we detail where our method stands in relation to current work in this area.

Graphical models provide a tractable way to depict concepts and the relationships between these concepts. However, there is a clear tug-of-war between model performance and transparency in this context \cite{holzinger2017we}. Consider, for example, that we build a simple Bayesian Model for predicting the severity of Diabetic Retinopathy, where each node in the Bayesian Model is a human-understandable concept, such as microanuerisms, dark spots, exudates, and hemorrhages.
Assuming we learn the structure and parameters of such a model, we would have a completely transparent technique for our task. However, it is difficult and computationally taxing to achieve the same level of performance with a Bayesian model as a deep neural network. This also requires an explicit differentiation or concept-level labelling of all relevant concepts expected to be in the Bayesian model, which is generally unavailable.

While Deep Neural Networks provide a much more efficient way to represent and learn from image data, they do not lend themselves to the simple conceptual analysis that graphical models like Bayesian Networks do. We propose a method to repurpose a trained deep learning model into an equivalent graphical structure at the level of abstract, human-understandable concepts. This provides us with a simple, transparent representation of the model's logic and allows us to determine the pathway it takes for making a prediction. Such a concept level representation is similar to that in deep probabilistic models, where the depth of the graph is considered over concepts instead of the depth of the computational graph \cite{goodfellow2016deep}.

We posit that such an abstraction is possible in a deep network since individual filters may be specialised to learn individual concepts. In the context of representation learning, it is hypothesized that deeper representation learning algorithms tend to discover more disentangled representations \cite{bengio2013deep}. For example, experiments in Network Dissection show that individual filters learn disentangled visual concepts \cite{bau2017network}. This behaviour has also been shown in the context of brain tumor segmentation models \cite{natekar2020demystifying}. 
Grouping filters which detect the same concept within a layer would then enable us to build a graphical representation of such concepts inherent in the network.  

This representation of the model has many advantages. It can tell us about the model's biases - for example, if it relies heavily on one concept for one class of predictions. It also allows us to determine active inference trails inherent in the model, as we have shown in this work. Our main contributions in this works are the following: (i) A method to represent a Deep Neural Network as a graphical model over abstract, high level concepts, encouraging concept-based explainability, and, (ii) Identification of inference trails from this graphical representation that help us understand the model's decision-making logic.

\section{Proposed Framework}

This work aims to abstract the model into an equivalent graphical model representation where concepts learnt by the network become nodes, and edges depict relationships between them. We take a clustering based approach to identify weights which may be detecting similar concepts in the input image. Such a method ensures that our explanations are independent of the input sample and that our formulations are computationally practical. Previous experiments show that for state-of-the-art DNNs trained on large-scale datasets like ImageNet \cite{imagenet_cvpr09}, euclidian distance in the activation space of final layers is an effective perceptual similarity metric \cite{zhang2018unreasonable}. It is not unreasonable that such behaviour extends to deep learning models in the medical domain. We use the euclidian distance between weight vectors averaged across the channel dimension as our similarity metric. 

\begin{figure*}[h]
 \centering
    \includegraphics[width=0.9\textwidth]{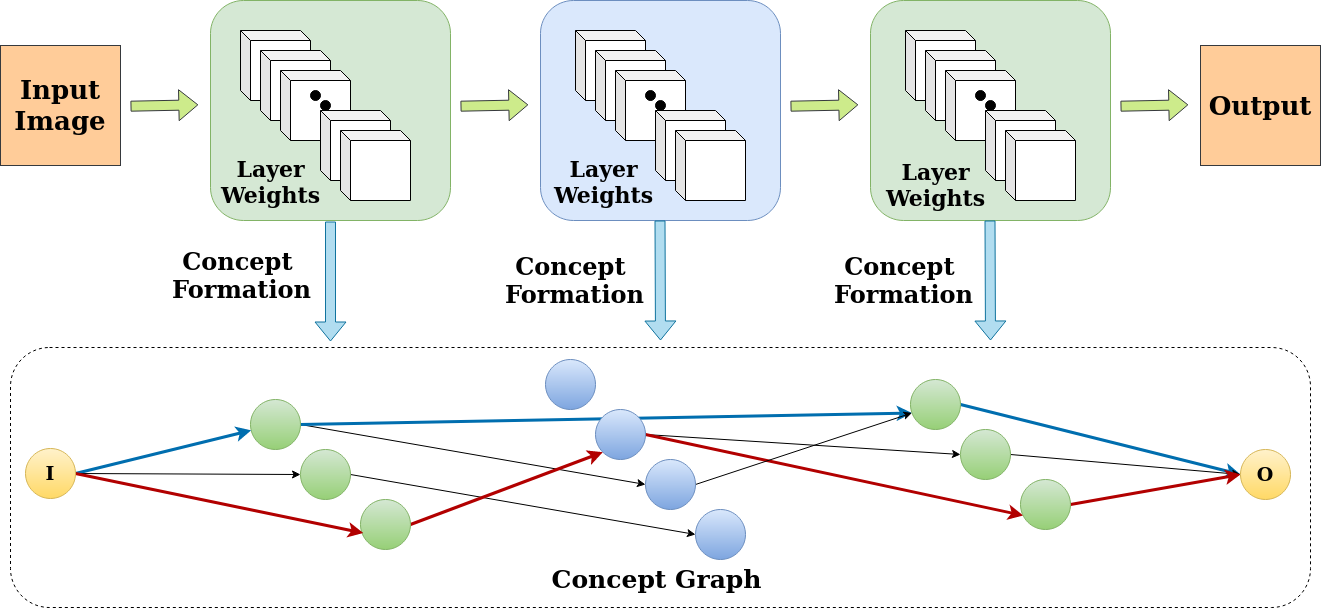}
    \caption{In the proposed framework, we construct a concept graph for a trained deep model. To generate concept graphs, we cluster weights in user-defined layers of the network, use them as concepts, and later estimate links based on a mutual information based metric. For example, trails represented in red and blue show active concept-level inference trails a network uses to predict the final result}
    \label{fig:framework}
\end{figure*}

We posit that the weight clusters thus identified are responsible for detecting individual concepts in the input image, and thus form the concept nodes in the abstracted graphical model. We visualize the concept detected by the clusters formed using a modification of Grad-CAM \cite{selvaraju2017grad}. Grad-CAM basically visualizes attention of a weight layer on the input image. By zeroing out weights from other clusters and only keeping weights from a particular cluster before obtaining Grad-CAM attention maps, we can find what the weight cluster corresponds to in the input space. 
Potential active inference trails are then found from the generated graphical model using a normalized mutual information based approach.

\begin{figure*}
 \centering
    \includegraphics[width=0.9\textwidth]{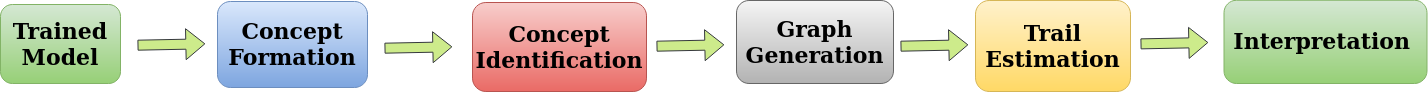}
    \caption{The above figure describes all the steps in the proposed concept-based interpretability framework visualized in Figure \ref{fig:framework}}
    \label{fig:pipeline}
\end{figure*}

The proposed framework for understanding the deep learning models consists of the following steps: (i) concept formation, (ii) concept identification, (iii) concept significance analysis, (iv) graph formation, and (v) trail estimation. Figure \ref{fig:framework} and \ref{fig:pipeline} provide a detailed overview of the described framework. Next, we go over each section of this framework in detail.

\subsection{Concept Formation}

We posit that groups of weight vectors in a layer are responsible for detecting a particular concept in the input image. Weight clustering has been used before in the context of network compression \cite{han2015deep,son2018clustering}. We show that a clustering based approach can be used to identify weights which are responsible for detecting a particular concept in the input image. Weight vectors can be clustered using a suitable metric and their attention over the input image can be used to determine the concept they are specialized to detect. Such an analysis can be performed at any level of granularity, for example one could perform the analysis choosing, say, only the first, fifth, ninth, and eleventh layers of a deep network so that a high level understanding can be gained of the concepts learnt by these layers.

Let the trained network be $\Phi(W, X)$, and the layers chosen for analysis be $\{..., l-n, l, l+m, ... \}$. The clusters $\{C^l_p, C^l_q, C^l_r, ...\}$ are formed as a result of clustering weights at layer $l$ in the network $\Phi$. Let $W = \{ w_1, w_2,..., w_n\}$ be the set of weights in a layer, where $W\in\mathbb{R}^{f \times f \times inc \times outc}$ and $w_i \in \mathbb{R}^{f \times f \times inc}$. Due to high dimensionality of the weight tensor, we take the mean of the weight tensor across the $outc$ dimension to obtain a representative tensor $w_i^{rep} = \frac{1}{inc} \sum_c w^c_i \in \mathbb{R}^{f \times f}$. To amplify the difference between symmetric weights we encode position information \cite{kori2018enhanced,palop2010quantifying} along with weights.

Clusters are formed using a hierarchical clustering method \cite{johnson1967hierarchical} using distance-based thresholding. This provides additional degrees of freedom to group weights into as many numbers of significantly different concepts. After obtaining the clusters, for visual verification we view the flattened weight vector to observe similarity among the clustered weights. Since direct visual interpretation is insufficient, to quantify the effectiveness of our clustering method we use $\mathbb{E}(SilhouetteScore)$ over all weights \cite{rousseeuw1987silhouettes} as a metric. Figure \ref{fig:clustering} in the Appendix \ref{appendix} depicts this for a sample layer. 

\subsection{Concept Identification}

In the Concept identification step, we try to associate formed weight clusters with some region in the input image which corresponds to a human-understandable \textit{concept}. 

Consider cluster $C^l_p$. To identify the concept learnt by the cluster and to depict this in a human understandable fashion, we first modify the trained network by dissecting the network at layer $l$, the outputs of which are denoted by $\Phi_l$. Then, we perform a variation of Grad-CAM (which we will simply refer to as concept attention maps), using the filters in the cluster $C^l_p$ as the outputs for which attention is to be computed, as described in equation \ref{eqn:relu}. 

In practice, this is done as follows. The dissected network $\Phi_l$ is modified by adding a $(1\times 1)$ convolution at the end, the weights of which are set to one. We then set the weights of all filters in the layer $l$ which do not belong to the cluster $p$ to zero. The effective operation performed by the added convolutional layer $\Phi_{l+1}$ is then equivalent to taking the mean across the channel dimension of only those filters which belong to the cluster, providing a single-channel condensation of the cluster which can be used for finding the concept-attention map. We denote the output of this layer by $\mathbb{E}_{k \sim idx_p} \Phi_{l,k}$, where $idx_p$ are the set indices in a layer $l$ belonging to cluster $C^l_p$, as formulated in equation \ref{eqn:gpool}.

Concept identification then amounts to finding the concept attention maps of this output with respect to the activations of the penultimate layer in the dissected network, i.e. $\Phi_{l-1}$ as described in equation \ref{eqn:grad}.

\begin{equation}
    y_p^l(x) = \frac{1}{Z} \sum_{i} \sum_{j} \Big( \mathbb{E}_{k \sim idx_p}
    \Phi_{l, k}(x) \Big)
    \label{eqn:gpool}
\end{equation}

\begin{equation}
    \beta_{m, p}^l(x) = \frac{1}{Z}\sum_{i} \sum_{j} \frac{\partial y_p^l(x)}{\partial \Phi_{l-1, m}(x)}
    \label{eqn:grad}
\end{equation}

\begin{equation}
    CAM_p^l = ReLU \left (  \sum_m \beta^l_{m,p}(x) \Phi_{l-1, m}(x) \right)  
    \label{eqn:relu}
\end{equation}

Where, $m$ is the index of a filter in layer $l-1$ and $k$ is index of filter in layer $l$, $\beta$ are the Grad-CAM importance weights, $i, j$ are the indices for the height and width dimensions of the feature map of the additional convolutional layer, and CAM is the output concept-attention map for concept $p$ of layer $l$. \\

\begin{figure*}[h!]
    \centering
    \subfloat[Layer: $19$, Gaussian Prior over entire weight layer]{\includegraphics[width=1.\textwidth]{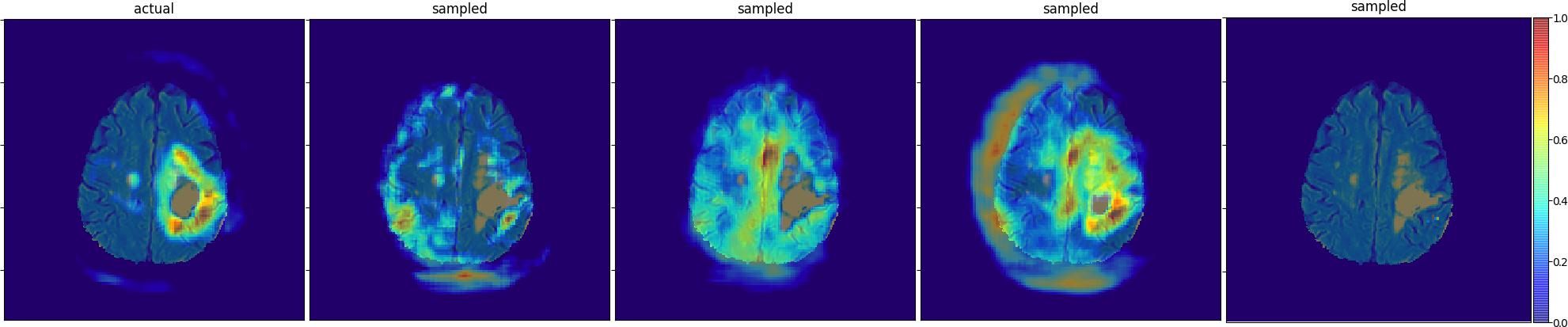}}\hfill
    \subfloat[Layer: $19$, Uniform Prior over only the weight cluster]{\includegraphics[width=1.\textwidth]{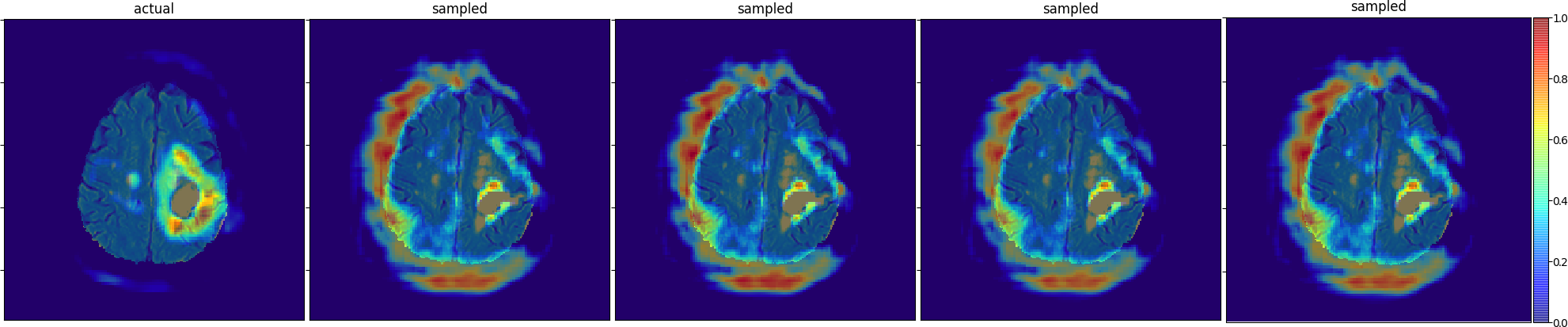}}\hfill
    \subfloat[Concept: $C^{19}_{2}$, Gaussian Prior over only the weight cluster ]{\includegraphics[width=1.\textwidth]{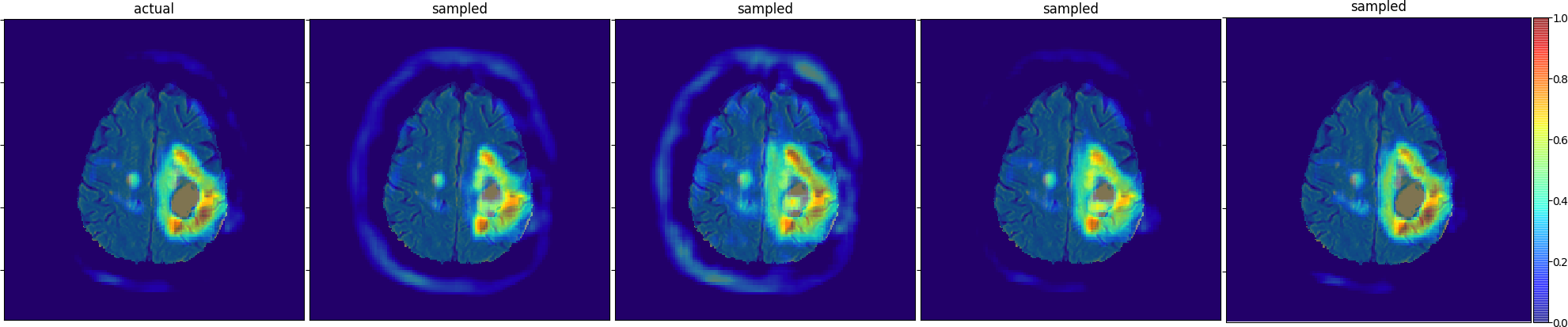}}\hfill
    \caption{This figure illustrates results of robustness experiments on BraTs data, (a) Concept attention maps by assuming Gaussian distribution over all the weights in a layer, (b) Concept attention maps by assuming Uniform distribution over only the cluster weights, and (c) Concept attention maps by assuming Gaussian distribution over only the cluster weights. Note that using a gaussian prior over only the cluster gives most consistent concept attention maps.}
    \label{fig:cluster_robustness_brats}
\end{figure*}

\begin{figure*}[h!]
    \centering
    \subfloat[Layer: $3d$ Gaussian Prior over entire weight layer]{\includegraphics[width=1.\textwidth]{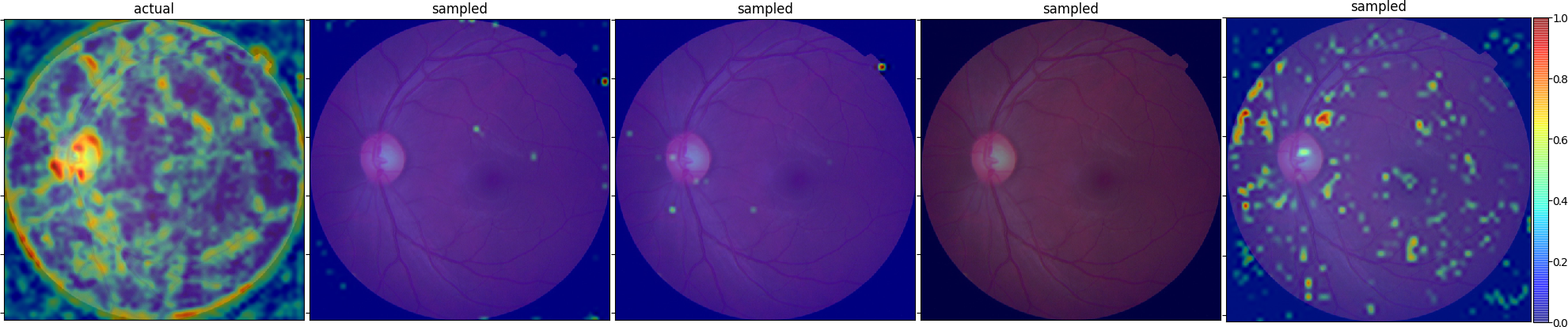}}\hfill
    \subfloat[Layer: $3d$ Uniform Prior over only the cluster weights ]{\includegraphics[width=1.\textwidth]{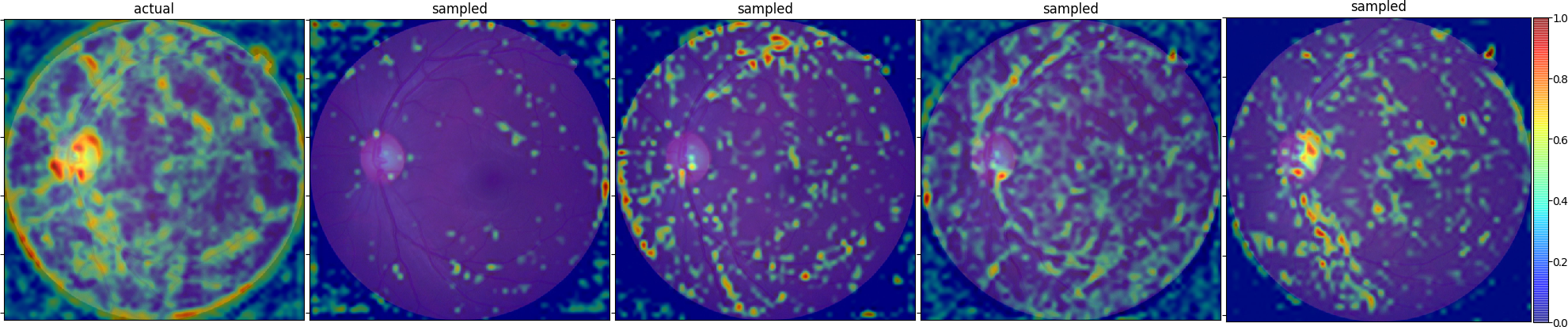}}\hfill
    \subfloat[Concept: $C^{3d}_{4}$ Gaussian Prior over only the cluster weights ]{\includegraphics[width=1.\textwidth]{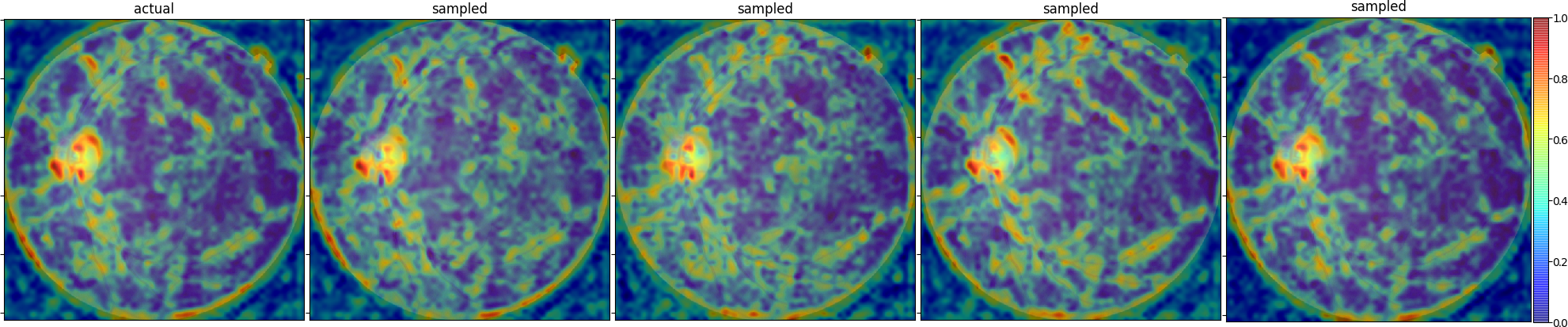}}\hfill
    \caption{This figure illustrates results of robustness experiments on APTOS data, (a) Concept attention maps by assuming Gaussian distribution over all the weights in a layer, (b) Concept attention maps by assuming Uniform distribution over only the cluster weights, and (c) Concept attention maps by assuming Gaussian distribution over only the cluster weights. Note that using a gaussian prior over only the cluster gives most consistent concept attention maps.}
    \label{fig:cluster_robustness_aptos}
\end{figure*}

\begin{figure*}[h!]
    \centering
    \subfloat[BraTS Concept: $C^{21}_{2}$ Tumor Core region ]{\includegraphics[width=1.\textwidth]{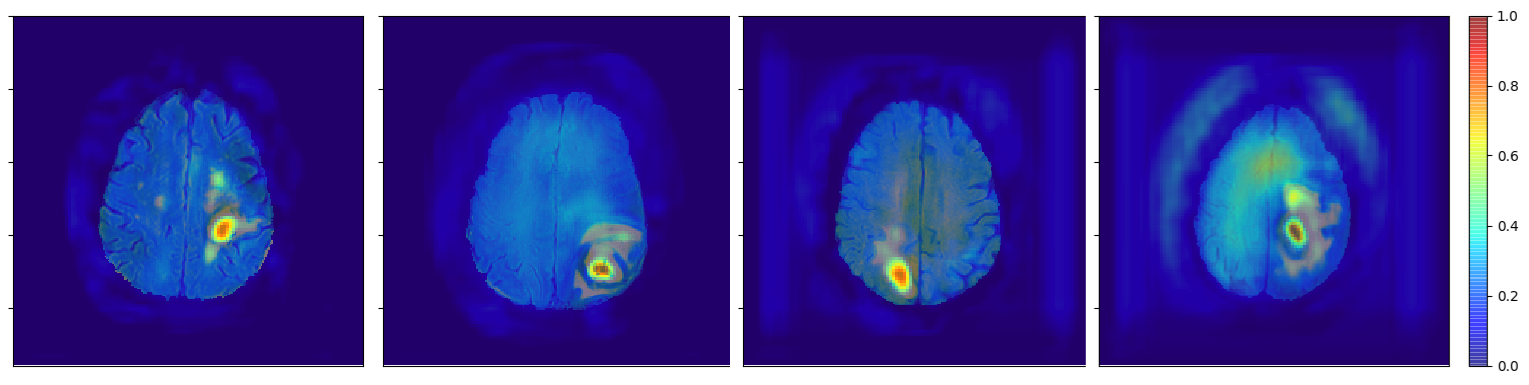}}\hfill
    \subfloat[BraTS Concept: $C^{19}_{2}$ Whole Tumor boundary]{\includegraphics[width=1.\textwidth]{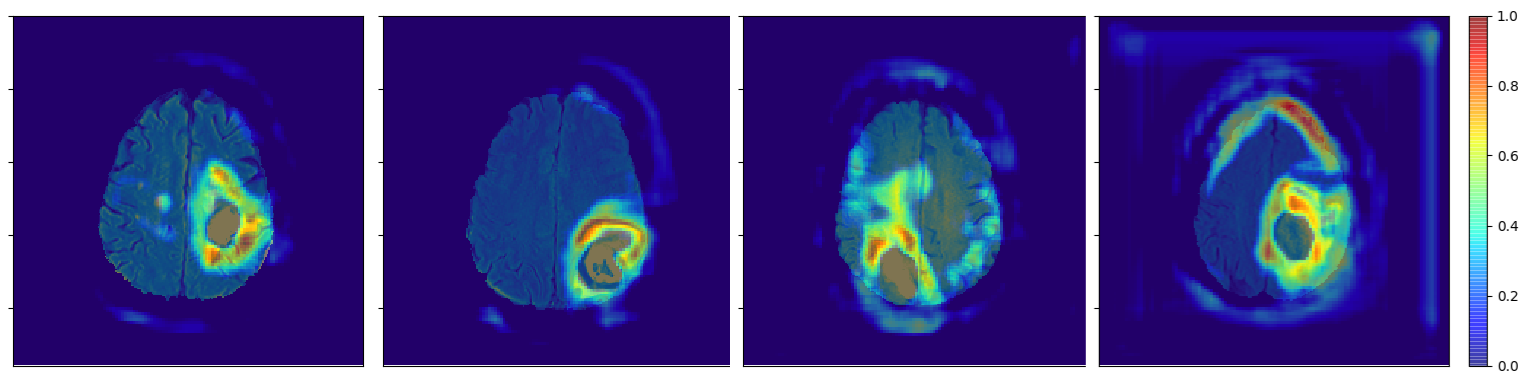}}\hfill
    \subfloat[APTOS Concept: $C^{2a}_{2}$ Lateral Eye boundary]{\includegraphics[width=1.\textwidth]{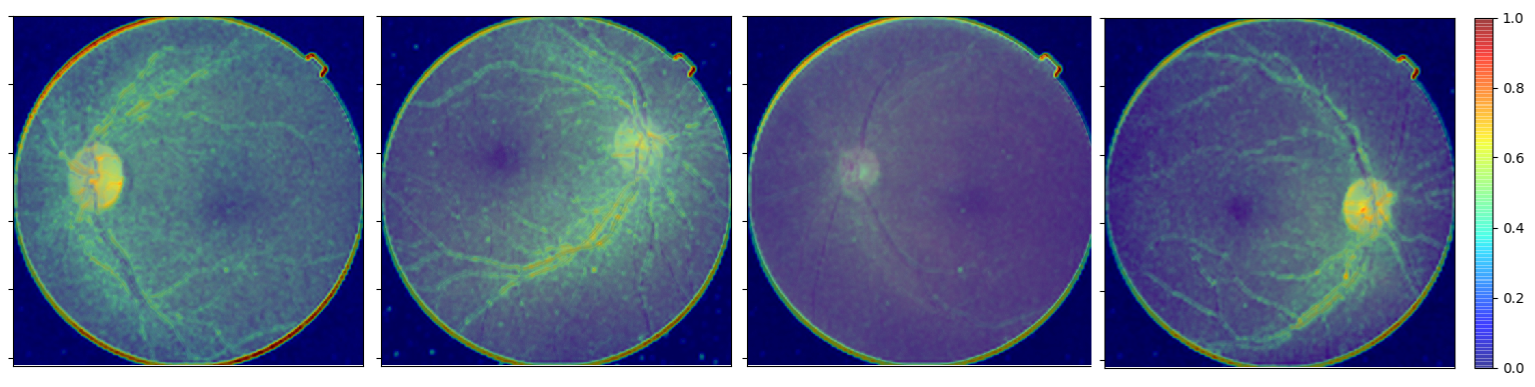}}\hfill
    \subfloat[APTOS Concept: $C^{3d}_{4}$ Major Blood vessels]{\includegraphics[width=1.\textwidth]{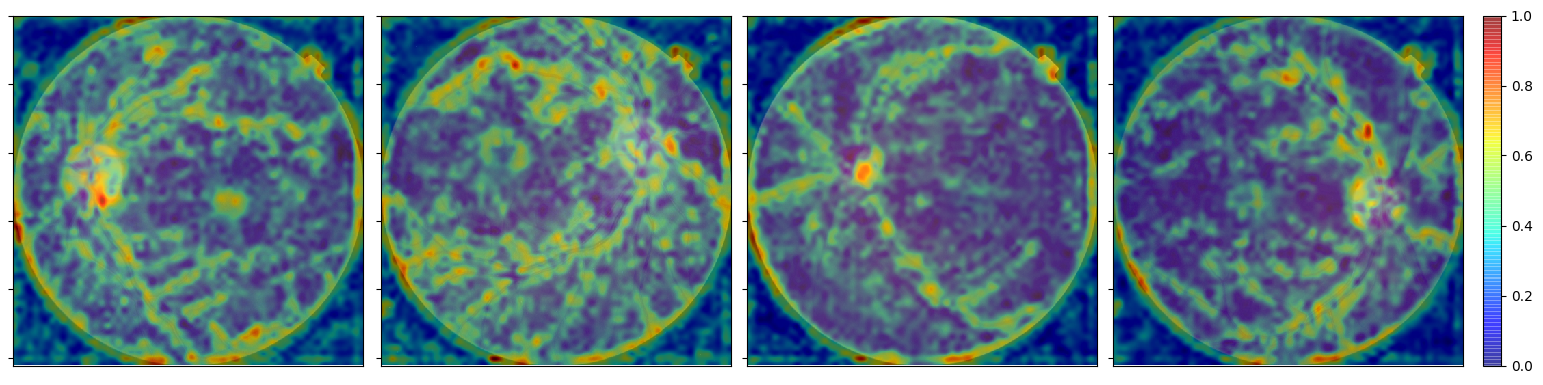}}\hfill
    \caption{The above figure shows the consistency of concept formation; each row indicates shows the concept-attention map for a cluster for different input samples}
    \label{fig:cluster_consistence}
\end{figure*}

Once the concepts are identified, we conduct significance tests to ensure that the concepts formed are consistent, robust, and localized. These procedures are detailed next. Figures \ref{fig:cluster_robustness_brats}, \ref{fig:cluster_robustness_aptos}, and \ref{fig:cluster_consistence} show the results of the conducted consistency and robustness tests for our identified concepts, which provide further evidence to support our hypothesis that groups of weight vectors in the model are responsible for detecting different semantic concepts. \\

\begin{figure*}[h!]
 \centering
    \includegraphics[width=0.9\textwidth]{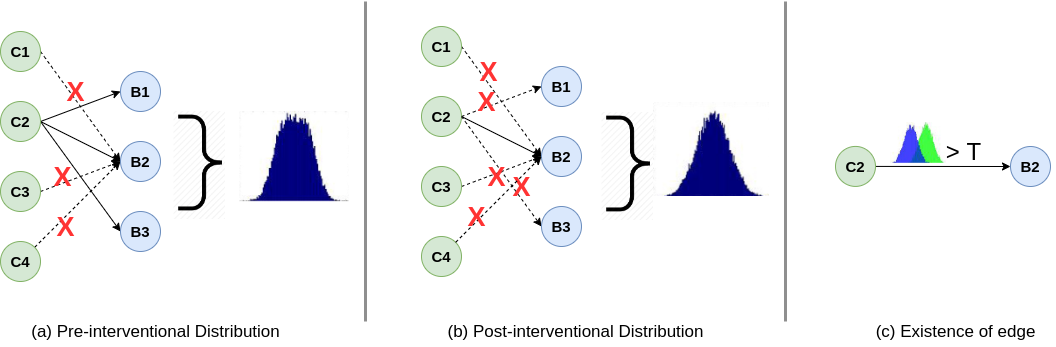}
    \caption{Above image describes the process of link formation. Sub-figure (a) describes how pre-interventional distribution is formed, sub-figure (b) describes how post-interventional distribution is formed, (c) exibits the condition for the existence of edge.}
    \label{fig:linkformation}
\end{figure*}

\textbf{Consistency: } To evaluate the consistency of clusters generated by the proposed method, we examine their regularity over multiple input samples in our datasets. Figure \ref{fig:cluster_consistence} illustrates the same, where each row corresponds to the concept attention map for an identified cluster over different images in the input dataset. It can be observed that identified clusters have similar concept attention maps for multiple input samples, irrespective of tumor location or optic disk location.\\

\textbf{Robustness: } Here, we try to evaluate the robustness of the formed clusters. Weights belonging to a specific layer in a neural network can be considered as i.i.d \cite{giryes2016deep}. We posit that after learning, all the weights belonging to a particular cluster come from an underlying distribution and are i.i.d. We assume a gaussian generating distribution for weights in the cluster and approximate this using the first and second order moment of the weights in the cluster. Figure \ref{fig:templateGraph} depicts this graphically.

Consider an identified cluster $C^l_p \in \mathbb{R}^{f \times f \times inc \times n}$, where $f$ is the filter size, $inc$ is the number of in-channels, and $n$ is the number of weights in the cluster. Let $w_i \in C^l_p$ be a weight belonging to the cluster $C^l_p$. Then, $w \in \mathbb{R}^{f \times f \times inc}$, i.e. the cluster $C^l_p$ contains $n$ weight tensors $w_i$ of size $f \times f \times inc$. We generate a gaussian distribution for each pixel $x_j$ at position $j$ in the flattened weight $w_i$, 

\begin{equation}
    x \sim \mathcal{N}(\mu, \sigma)
    \label{eqn:dist}
\end{equation}

\begin{equation}
    \mu = \mathbb{E}_{i}(x_j), \sigma = \mathbb{E}_{i}(x_j - \mathbb{E}_{i}(x_j))
    \label{eqn:dist1}
\end{equation}

We then sample $n$ number of weights as detailed above, replace all $n$ weights in the cluster $C^l_p$ by the sampled weights, and recompute our concept attention maps. Figures \ref{fig:cluster_robustness_brats} and \ref{fig:cluster_robustness_aptos} show the results of this experiment.

We observe that recomputed concept attention maps correspond to the same region in the input space as the original concept attention maps. We also generate recomputed concept attention maps using a uniform prior over the cluster weights as well as a gaussian prior taken over the range of all weights in the layer, and compare this with the results of using a gaussian prior over only the cluster weights. 
It can be observed that concept attention maps (CAMs) formed by using gaussian priors over only the weights belonging to that particular cluster are visually similar to the originals for each sampling run, while CAMs formed using uniform priors or CAMs formed using gaussian priors over all the weights do not encode the same concept in the input space and show high variability for each sampling run. This behaviour is seen consistently over all input samples. Thus, we empirically justify that our identified concepts come from the same underlying distribution, and that the gaussian is a reasonable proxy for this distribution.


\subsection{Network Formation and Information Flow}

\begin{figure*}[h!]
    \centering
    \includegraphics[width=0.85\textwidth]{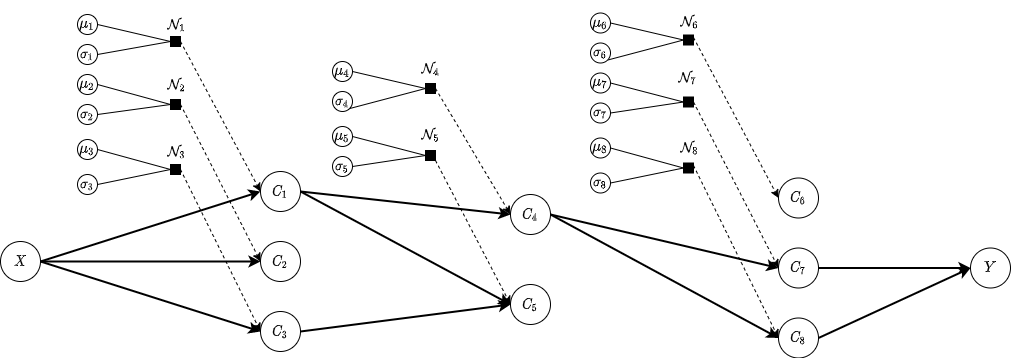}
    \caption{A visual depiction of the constructed graphical representation for the network given the set of layers to analyse. Each pixel in a concept can be imagined to be drawn from its own gaussian distribution, using the mean and variance of the pixel over the cluster as parameters. Dotted arrows show the concept is sampled from its corresponding normal distribution. Dark arrows show links between concepts.}
    \label{fig:templateGraph}
\end{figure*}

\begin{figure*}[h!]
    \centering
    \includegraphics[width=\textwidth]{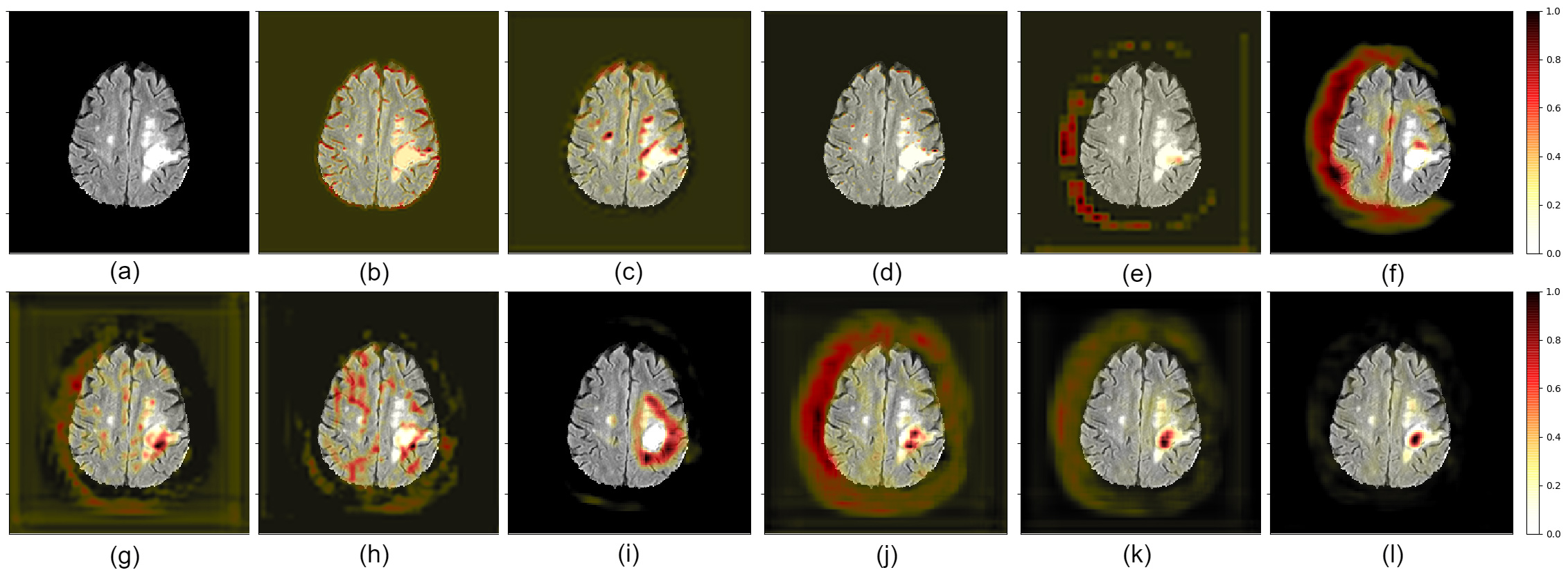}
    \caption{%
    Concepts obtained from various layers of a trained U-net model superposed over the MRI Flair channel.
    (a) $C^3_0$: doesn't capture any input region,
    (b) $C^3_1$: concave edges,
    (c) $C^3_2$: linear edges,
    (d) $C^5_2$: interior key points.
    (e) $C^{13}_0$: Lateral left hemispherical brain boundary,
    (f) $C^{13}_3$: Lateral left hemispherical and tumor core brain boundary,  
    (g) $C^{15}_2$: Anterior tumor boundary,
    (h) $C^{15}_3$: Tumor core boundary,
    (i) $C^{19}_2$: Whole tumor boundary,
    (j) $C^{17}_0$: Lateral brain boundary and tumor core boundary,
    (k) $C^{21}_1$: Diffused tumor core region,
    (l) $C^{21}_2$: Tumor core region.
    }
    \label{fig:brainConcepts}
\end{figure*}

Once concepts and have been identified for the given set of layers, we have the means to construct our equivalent graphical representation. 

Given these concepts, we can identify relationships between them to generate a human-understandable trace of inference which augments model predictions. In order to identify the relationship between two concepts, we compute the normalized mutual information between the pre-interventional and post-interventional feature map distribution, as described below. 


For the directed link between two concepts in layer $p$ and $q$, $C_i^p \rightarrow C_j^q$, the pre-interventional distribution $\mathbb{P}(\Phi_j(x ~|~ do(C_{-i}^p = 0)))$ is the feature map distribution obtained on zeroing out the weights belonging to all concepts other than $C_i^p$ in layer $p$ (i.e., $do(C_{-i}^p = 0)$, where the $do$ operator indicates a manual intervention on the argument to set it to a particular value, which is 0 in this case). This distribution tells us about information flowing from $C_i^p$ to all concepts in the succeeding layer $q$. Similarly, the post-interventional distribution $\mathbb{Q}(\Phi_j(x ~|~ do(C_{-i}^p = 0), do(C_{-j}^q = 0)))$ is the feature map distribution obtained at the layer $q$ by zeroing out the weights belonging to all the clusters other than $i$ in layer $p$ as well as the weights belonging to all the clusters other than $j$ in layer $q$ (i.e., $do(C_{-i}^p = 0)$ and $do(C_{-j}^q = 0)$). This distribution tells us about the information flowing only from $C_i^p$ to $C_j^q$. In this formulation the terms \textit{pre} and \textit{post} interventional are considered only with respect to layer $q$. Figure \ref{fig:linkformation} shows this process graphically.

Based on our formulation, the directed link $C_i^p \rightarrow C_j^q$, exists only if equation \ref{eqn:nmi} is satisfied. 

\begin{equation}
\begin{split}
    \text{NMI}\big( \mathbb{Q} \big(\Phi_j(x ~|~ do(C_{-i}^p = 0), do(C_{-j}^q = 0))\big), \\
    \mathbb{P}\big(\Phi_j(x ~|~ do(C_{-i}^p = 0))\big)\big) > T
    \label{eqn:nmi}
\end{split}
\end{equation}

This basically states that the link exists only if the mutual information between pre and post interventional distribution is higher than a set threshold. High mutual information implies that a significant portion of the information flowing from the concept $C_i^p$ to layer $q$ occurs through that specific link $C_i^p \rightarrow C_j^q$. This results in the formation of a concept graph, an example visualization of which is shown in Figure \ref{fig:templateGraph}. Note that this graphical model is not intended to be complete, only representative. Since our graph can be constructed over any set of layers chosen by the user, there could be multiple inference trails that denote relationships between different concepts.

\subsection{Trail Estimation}

Given our graphical representation and the existence of links between concepts, we now have a method to track inference steps taken by the model. The obtained concept graph is a DAG with depth $m$, where $m$ is number of layers specified by the user for interpretability. The trails are all the paths running from input to a particular node used in an inference. The obtained trails encode the flow of concept level information used in making a prediction.

For example, consider the sample trail $X \rightarrow C_1 \rightarrow C_4 \rightarrow C_8 \rightarrow Y$ in Figure \ref{fig:templateGraph}. Medical professionals can then highlight whether or not such an inference trail makes sense from a biomedical perspective, and understand the model's biases and its common logical steps of inference. The next section details the application of the above framework on benchmark biomedical image datasets.


\section{Experiments}

We illustrate the working of our proposed framework on both classification and segmentation tasks. For the classification task, we considered the Diabetic Retinopathy problem, and for segmentation, we considered the Brain Tumor Segmentation problem. In both the experiments, the aim was to explain the building blocks of the model, and understand the hierarchy of decision making in deep learning models. All the experiments and results can be reproduced by using notebooks provided in the code repository  \url{https://github.com/koriavinash1/BioExp_Experiments}.

\subsection{Brain Tumor Segmentation}

In the past decade, there has been significant development of image processing algorithms for segmenting intra-tumoral structures in brain MRI images \cite{bakas2018identifying}. Deep Learning has shown great potential in this context, with the BraTS challenge \cite{kamnitsas2017efficient,wang2017automatic,myronenko20183d,kori2018ensemble} setting the benchmark for research in this area. The BraTS dataset contains nearly 300 brain MRI volumes annotated by experts for tumor regions. Various deep learning algorithms have shown great performance in segmenting tumor core, enhancing tumor, and edema regions from these MRI volumes. 

We implement our algorithm on a UNet based model for brain tumor segmentation, which is a popular segmentation architecture in the medical context \cite{ronneberger2015u}. Our model also has residual connections as per \cite{kermi2018deep}, and achieves a dice score of 0.788, 0.743, and 0.649 on whole tumor, tumor core and enhancing tumor segmentation respectively on a held-out validation set of 48 volumes. Our model is not meant to achieve state of the art performance. Instead, we aim to demonstrate our method on a commonly used architecture for brain-tumor segmentation. The next sections detail the concepts and active inference trails obtained as a result of our framework on this task.

\begin{figure*}[h!]
    \centering
    {\includegraphics[width=1.\textwidth]{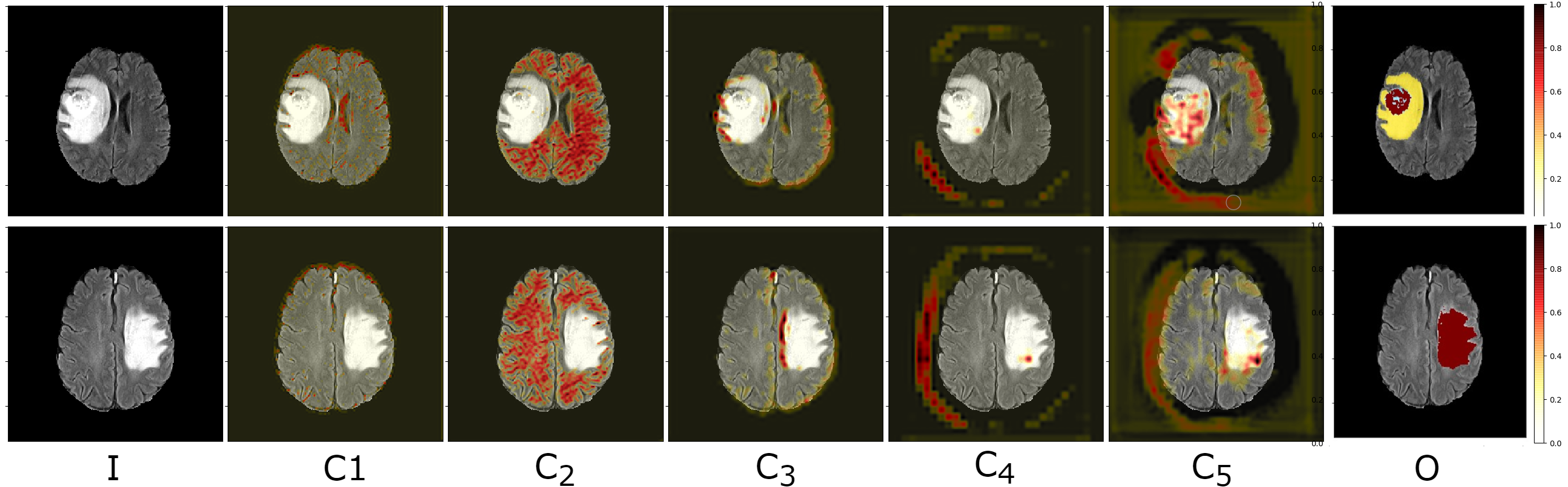}}\hfill
    \caption{Active inference trail for enhancing tumor (Each row is a trail for one input sample, red regions are high attention): \small{\textit{(\textbf{I}: Input image to a network)  $->$  (\textbf{$C_1$}: Concave edges)  $->$  (\textbf{$C_2$}: White matter region)  $->$  (\textbf{$C_3$}: Tumor boundary)  $->$  \textbf{$C_4$}: (Lateral brain boundary)  $->$  (\textbf{$C_5$}: Inferior tumor boundary)  $->$  (Enhancing Tumor)}}}
    \label{fig:brats_trail}
\end{figure*}

\begin{figure*}[h!]
    \centering
    {\includegraphics[width=1.\textwidth]{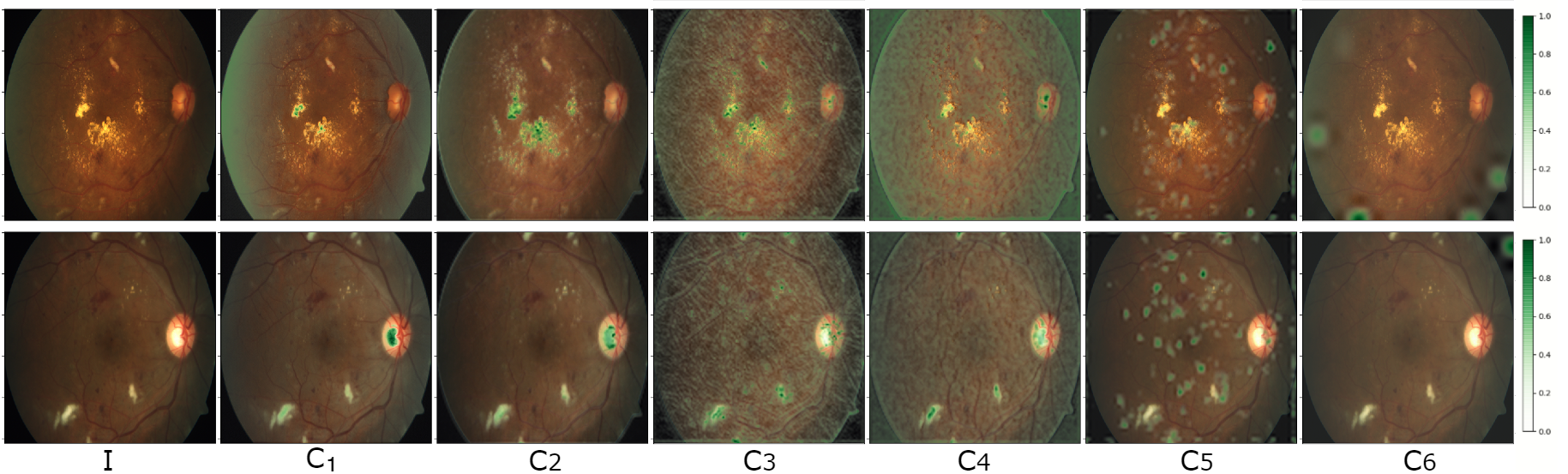}}\hfill
    \caption{Active inference trail for severe DR (green regions are high attention): \textit{(I: Input Image)  $->$  ($C_1$: Optic Cup/Hard exudates)  $->$  ($C_2$: Hard Exudates)  $->$  ($C_3$: Blood vessels, soft exudates)  $->$  ($C_4$: Blood vessel, soft exudates)  $->$  ($C_5$: dot-blot Hemorrhages/laser scar marks of retinal photocoagulation)}}
    \label{fig:aptos_trail}
\end{figure*}
\vspace{-1.5mm}
\subsubsection{Concepts}
The  $\mathbb{E}(SilhouetteScore)$ over all the data-points is $0.241$, indicating the formation of weak but significant clusters. Figure \ref{fig:brainConcepts} describes the various concepts identified from our model. Initial layers (convolutional layers 3 and 5) correspond to edges in a specific direction or brain boundaries. In higher layers, filters start capturing more local information. It can be observed that some concepts capture brain boundary, while some capture tumor boundary. Figure \ref{fig:brainConcepts} contains a description of the various concepts obtained from out network. This behaviour is in line with the understanding that filters in shallower layers of brain tumor segmentation models learn simple patterns while deeper layers learn progressively more complex concepts \cite{natekar2020demystifying}. The brain atlas described in \cite{ding2016comprehensive} was used to formulate appropriate descriptions.

\subsubsection{Trails and Discoveries} \label{brats_trails}
 Figure \ref{fig:brats_trail} describes inference trails involved in predicting the enhancing tumor region (Trails for other classes are available in the Appendix \ref{appendix}). These show the model's attention is initially on the outer edges and keypoints of the brain, then moves to the white and grey matter region, then the tumor boundary, and finally the internal tumor region. The caption of Figure \ref{fig:brats_trail} also provides a description of the visual trails for an image based on the predefined concept description. In the discussion section, we analyse these trails with feedback from a certified radiologist.

\subsection{Diabetic Retinopathy classification}

Diabetic Retinopathy (DR) is frequent in individuals suffering from diabetes \cite{fong2004retinopathy}. Deep Learning algorithms have shown great promise in detecting the severity of diabetic retinopathy and have the potential to greatly simplify diagnosis and detection. We implement our framework on a ResNet50 based network which achieves a Cohen Kappa Score of 0.71 on the validation set of the APTOS dataset \cite{aptosdataset}. The APTOS dataset contains around 5500 retina images taken using fundus photography. The severity of diabetic retinopathy has been rated for each image on a scale of 0 (no DR) to 4 (Proliferative DR). 
Each stage of DR is characterized by certain features - such as microanuerisms, exudates, and hemorrhages. Thus, it becomes necessary to see whether deep learning models process and identify these features, and to see the model's understanding of relationships between these and the predicted severity of DR. We follow a similar process as that for brain tumor segmentation, detailed below.


\subsubsection{Concepts}
\begin{figure*}[h!]
    \centering
    \includegraphics[width=\textwidth]{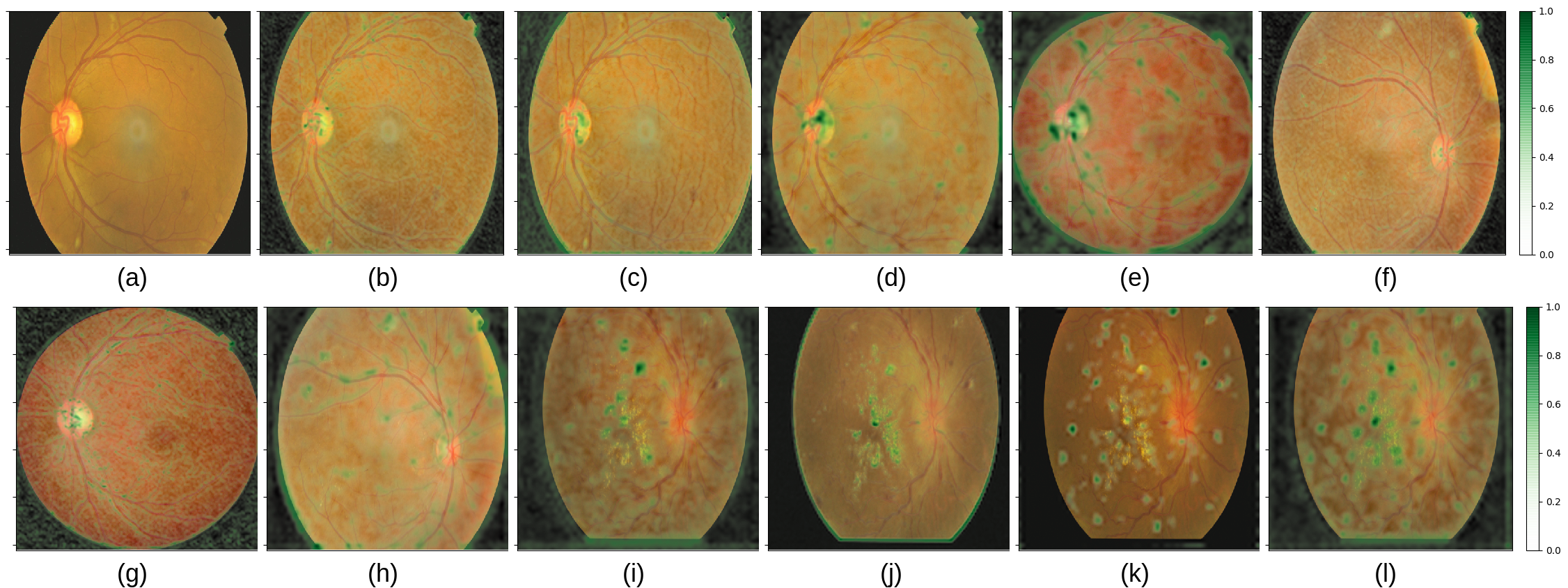}
    \caption{%
    This figure illustrates the concepts obtained from various layers of a trained ResNet50 model. Based on the region of activation we provide description of the concepts as follows: 
    (a) $C^{1}_{1} $ : doesn't capture any input region,
    (b) $C^{1}_{2} $ : Right lateral edges,
    (c) $C^{2a}_{1} $: Lateral edges,
    (d) $C^{2a}_{2} $: Optic disk + lateral edges, 
    (e) $C^{2c}_{2} $: Optic disk + blood vessels,
    (f) $C^{3a}_{2} $: All blood vessels (tiny),
    (g) $C^{3d}_{4} $: Major blood vessels,
    (h) $C^{3d}_{5} $: Blood vessels (eroded),
    (i) $C^{4a}_{2} $: Yellow spots (may be hard exodates),
    (j) $C^{4f}_{1} $: Yellow spots (may be hard exodates),
    (k) $C^{4a}_{3} $: Pale Yellow (may be hard exodates),
    (l) $C^{5c}_{2} $: Hard/Soft exodates
    }
    \label{fig:aptosConcepts}
\end{figure*}

The  $\mathbb{E}(SilhouetteScore)$ over all the data-points is $0.2$, which again indicates the formation of weak but significant clusters. Figure \ref{fig:aptosConcepts} describes the identified local and global level concepts, encoding blood vessels, hard and soft exudates, dot-blot hemorrhages, etc. 

\subsubsection{Trails and Discoveries}
Similar to the trails obtained for the BraTS dataset, we show example inference trails obtained for the APTOS dataset in Figure \ref{fig:aptos_trail} and Figures \ref{fig:aptos_trail_sup} and \ref{fig:aptos_trail_sup_2} in the Appendix. These describe visual trails involved in predicting 'Severe', 'Moderate', and  'Proliferative' classes of diabetic retinopathy respectively. An ophthalmologist's feedback was obtained on the concept trails, which is elaborated in the discussion section. Once again, we see the emergence of medically relevant concepts in a hierarchical manner, which may provide additional support to medical professionals apart from just the output classification.

\section{Related Work}
\label{section: related}

Explainability is generally categorized into post-hoc and ante-hoc methods, where post-hoc explainability methods try to analyze and make inferences on trained models \cite{simonyan2013deep,zeiler2014visualizing,ustun2014methods}. In contrast, ante-hoc methods try to build an explainable model while training itself \cite{caruana2015intelligible,holzinger2017glass,holzinger2019interactive}.

Current research directions in post-hoc interpretability focus mainly on visualizing network attributions or illustrative samples in the input space \cite{selvaraju2017grad,bau2017network,olah2017feature,kim2018interpretability}. Our work is related to methods involving disentangled latent representations and concept based explanations. For example, previous experiments on network dissection show that deep networks learn disentangled latent concepts \cite{bau2017network}. Previous concept based interpretability methods \cite{ghorbani2019towards,kim2018interpretability} use input patches to identify salient concepts that lead to a particular output. This has been extended to include a completeness measure for identified concepts \cite{yeh2019concept}. However, neither of these methods consider the relationship between concepts learnt by the model and do not provide a trace of inference steps. Also, these methods either require a pre-processed set of input samples as concepts \cite{kim2018interpretability}, or automatically segment the input image at various resolutions to create concepts \cite{ghorbani2019towards}. However, in the medical domain, obtaining such concepts is difficult - manual concept curation is time consuming and would require medical experts, while segmenting the input image may not lead to the formation of coherent anatomical concepts which add interpretability value, especially in cases where the task itself is image segmentation. In such domains, interpretability needs to emerge organically from the model itself and provide an understanding of the model's decision making logic. 

Our work introduces a post-hoc interpretability method, by abstracting the trained model into interpretable \textit{concept graphs}, where concepts and their relationships emerge implicitly from the model, doing away with the need for user-curated input concepts. Our concept graphs allow easy visualization of the model's logic on an abstract, human-understandable level.

\section{Discussion}
\label{sec:discussion}
This work aims to provide concept-based interpretability for deep neural networks, demonstrating the results on medical data. We use a clustering technique to extract a graphical representation of concepts in the network, and visualize the clustered concepts using a variation of Grad-CAM.  We then use an information-theoretic measure to determine relationships between concepts and build concept level inference trails within our network. Our results show that consistent, distinct trails that lead to a particular classification made up of anatomically relevant concepts can be identified.

While in previous work on interpretability in the medical domain \cite{natekar2020demystifying}, the existence of disentangled concepts is shown in brain-tumor segmentation networks, in this work we create a concept-level graph that depicts the relationships between these concepts and provides an understanding of inference trails in the model. As opposed to previous concept-based approaches \cite{ghorbani2019towards,kim2018interpretability}, no manual extraction of concepts from the input dataset is required, which is a challenging task in the medical domain. In this initial work, we demonstrate the potential of our technique on two medical datasets - the BraTS dataset for brain tumor segmentation and the APTOS dataset for diabetic retinopathy classification. 

For brain-tumor segmentation, a certified radiologist's comments on the extracted concept trail was solicited. They noted the lateral to medial and anterior to superior nature of attention of the model, as well as the hierarchical approach to segmentation which is in line with a radiologist's thought process. They commented that tumour boundary delineation as seen in Figure \ref{fig:brats_trail} concept $C_3$ has value for neurosurgeons when obtaining biopsy or resecting the tumour since this helps prevent damage to unaffected brain tissue. They also noted that a neuroradiologist would be able to immediately perceive the presence of gliomas in the flair sequence and it is in general not possible to break down that perception in terms of the trails obtained from the concept graphs. However, the visualization of concepts that are focused on tumour boundaries and the tumour core would help in improving confidence and trust in the deep learning model. The tumor core and characteristics are also defined which will aid in diagnosis and grading of the tumor. 

For Diabetic Retinopathy, an ophthalmologist's feedback was obtained on the output trail described in Figure \ref{fig:aptos_trail}. Various features, such as hard and soft exudates, dot-blot haemorrhages, optic cup, and laser scar marks of retinal photo-coagulation were identified. In the case of DR, it is interesting that features like this, which ophthalmologists look at to classify DR images, emerge implicitly from the model, even though it has not been explicitly trained to learn these.


\subsubsection{Acknowledgments.}
We would like to acknowledge help from Dr. Ravikanth Balaji and Dr. Devika Joshi for providing clinician (radiological and opthalmological) feedback on the inference trails obtained.

\bibliographystyle{aaai}
\bibliography{main}
\clearpage 

\onecolumn
\section{Appendix I}\label{appendix}

\vspace{4mm}
Here we show additional figures and examples which result from our primary analysis above. First, the results for cluster significance tests are shown - robustness and consistency. Then we show additional examples for brain-tumor segmentation and diabetic retinopathy classification, as well as other supporting images.

\begin{figure*}[h!]
 \centering
    \includegraphics[width=\textwidth]{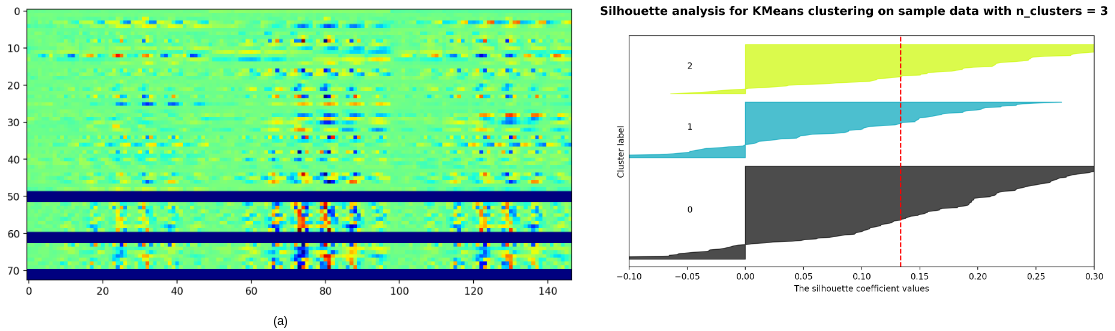}
    \caption{Above image describes the effectiveness of clustering. Sub-figure (a) describes the initial layer weights from ResNet50 trained on APTOS \cite{aptosdataset} data, in the figure dark blue horizontal bands seperates the weights among multiple clusters (provided figure has 3 clusters). Sub-figure (b) quantifies the effectiveness of clusters obtained as the result of proposed method using a silhouette plot}
    \label{fig:clustering}
\end{figure*}

\begin{figure*}[h!]
    \centering{\includegraphics[width=1.\textwidth]{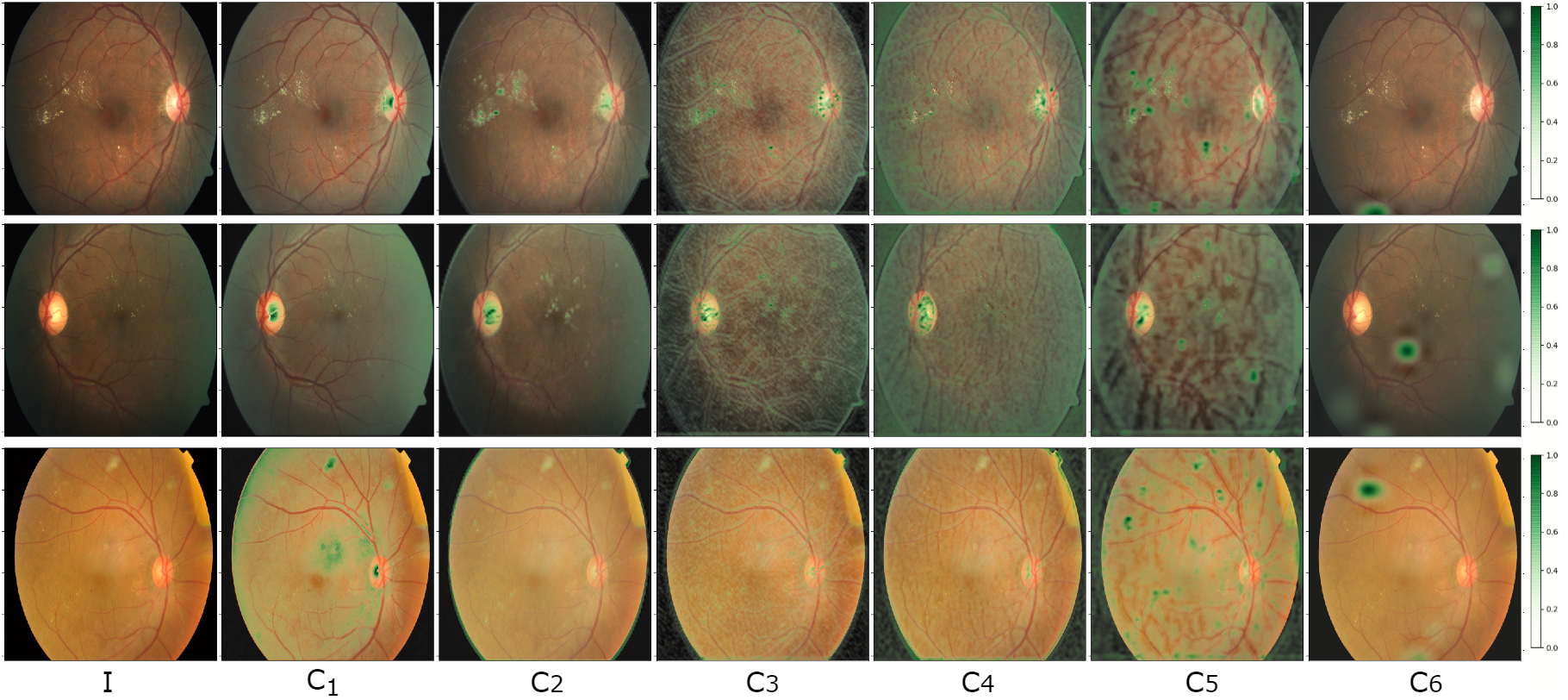}}\hfill
   \caption{Active inference trail for Moderate DR (Green regions are high attention): \textit{(I: Input Image to a network)  $->$  ($C_1$: Soft exudates + Optic Cup)  $->$  ($C_2$: Hard exudates)  $->$  ($C_3$: All blood vessels)  $->$  ($C_4$: Optic disk and blood vessels)  $->$  ($C_5$: Inverted Blood vessel (eroded) Image)  $->$  ($C_6$: Dark spots)}}
    \label{fig:aptos_trail_sup}
\end{figure*}

\begin{figure*}[h!]
    \centering{\includegraphics[width=1.\textwidth]{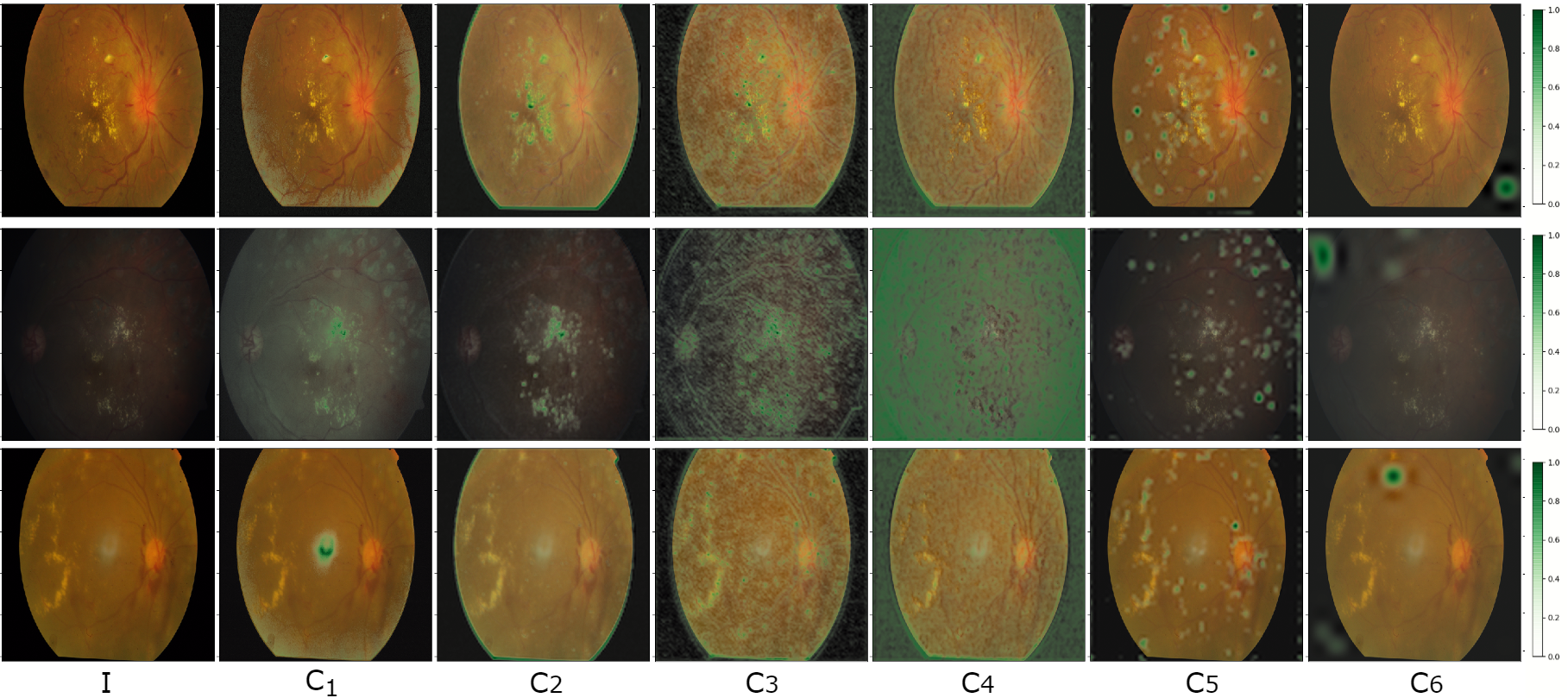}}\hfill
   \caption{Active inference trail for Proliferative DR (Green regions are high attention): \textit{(I: Input Image to a network)  $->$  ($C_1$: Pale areas, due to attenuated artery endings + macula) $->$  ($C_2$: Hard exudates)  $->$  ($C_3$: All blood vessels + key points)  $->$  ($C_4$: Optic disk and blood vessels)  $->$  ($C_5$: Laser scar marks of retinal photocoagulation + blot haemorrhages  $->$  ($C_6$: Dark spots)}}
    \label{fig:aptos_trail_sup_2}
\end{figure*}

\begin{figure*}[h!]
    \centering
    {\includegraphics[width=1.\textwidth]{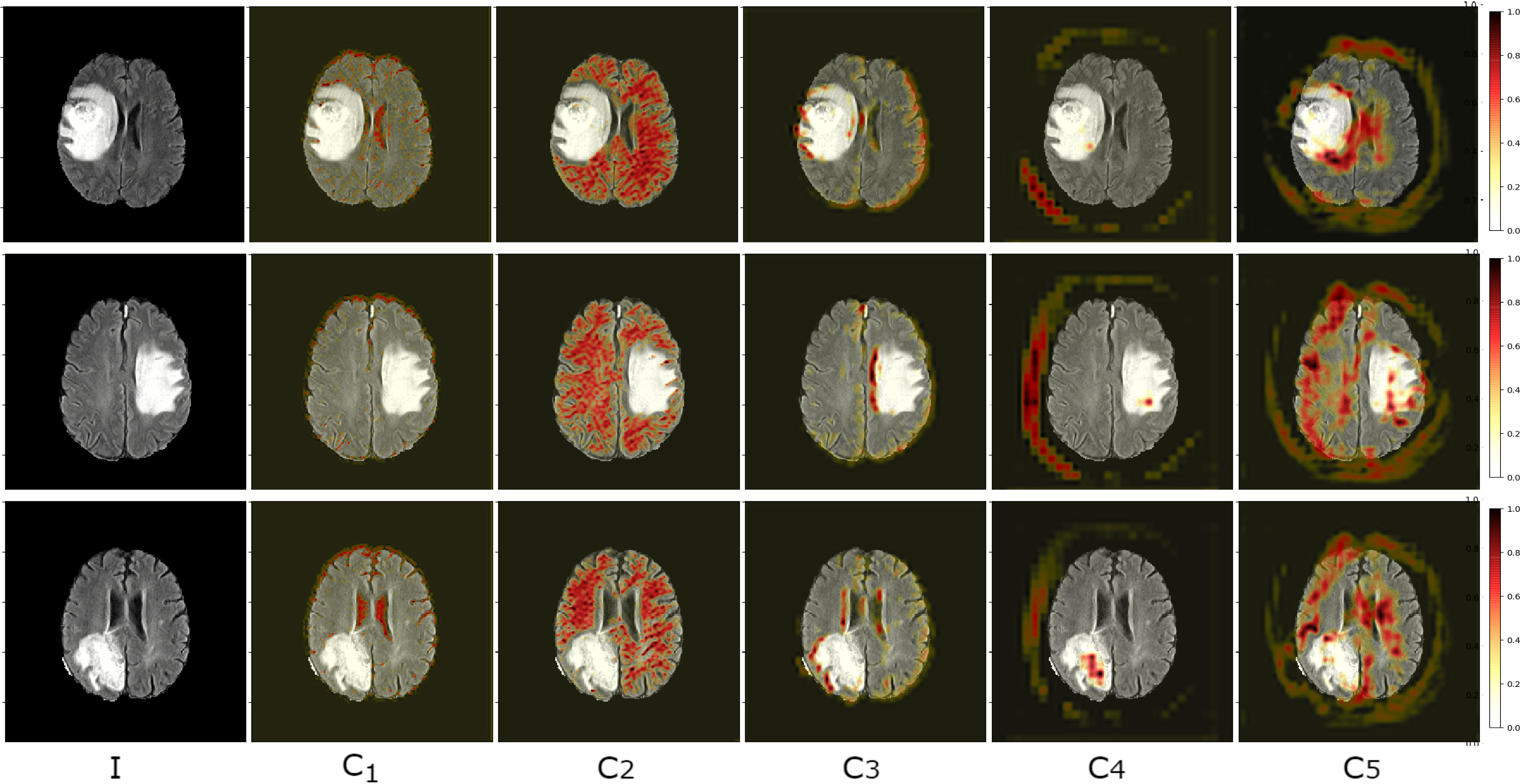}}\hfill
    \caption{Active inference trail for Edema (Each row is a trail for one input sample, red regions are high attention): \small{\textit{(\textbf{I}: Input Image to a network)  $->$  (\textbf{$C_1$}: Concave edges)  $->$  (\textbf{$C_2$}: White matter)  $->$  (\textbf{$C_3$}: Brain and tumor boundary)  $->$  \textbf{$C_4$}: (Lateral brain boundary)  $->$  (\textbf{$C_5$}: Lateral tumor boundary and mid brain)  $->$  (Edema region)}}}
    \label{fig:brats_trail_sup}
\end{figure*}

\begin{figure*}[h!]
    \centering
    {\includegraphics[width=1.\textwidth]{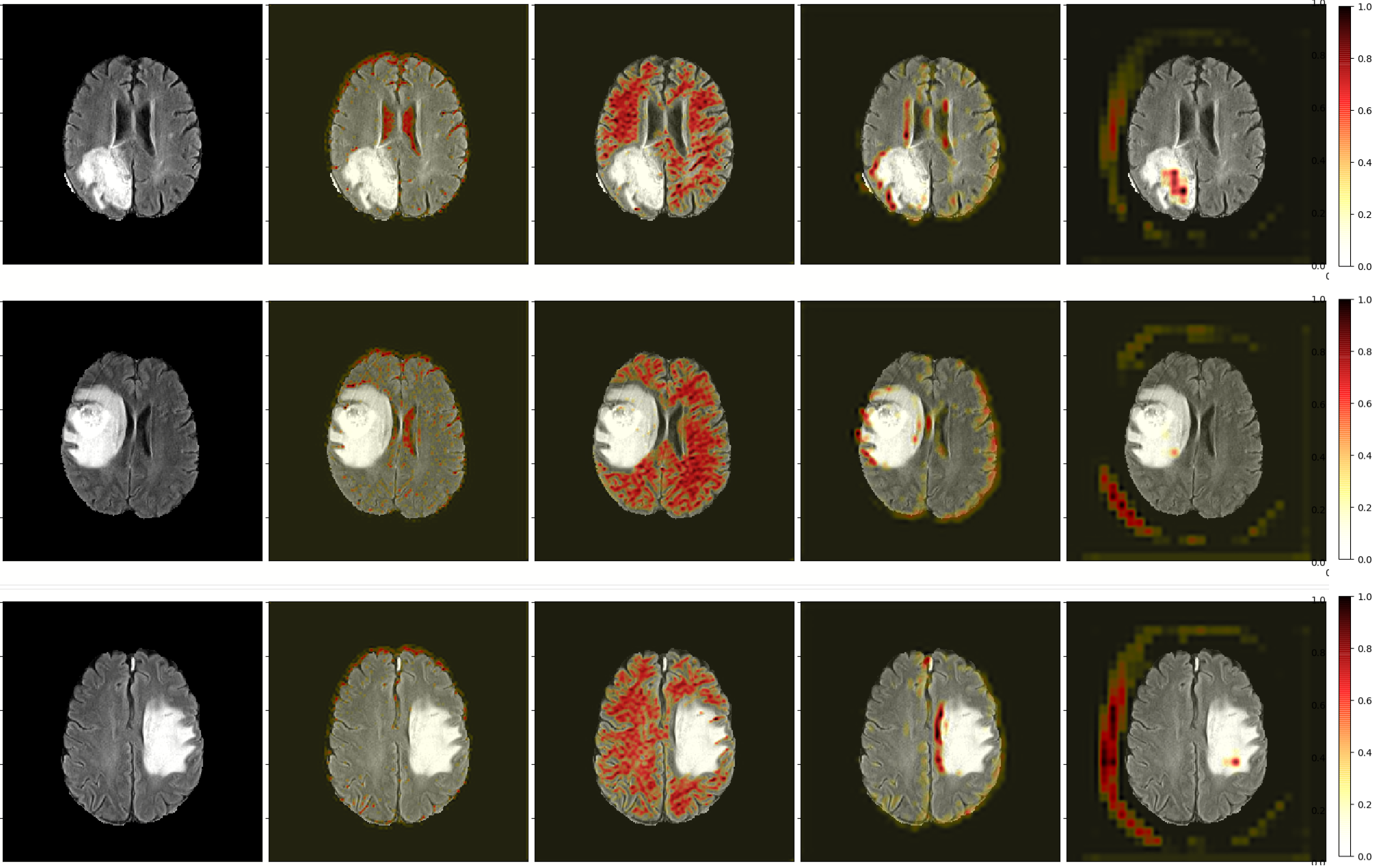}}\hfill
    \caption{Active inference trail for Tumor Core (Each row is a trail for one input sample, red regions are high attention): \small{\textit{(\textbf{I}: Input Image to a network)  $->$  (\textbf{$C_1$}: Concave edges)  $->$  (\textbf{$C_2$}: White matter)  $->$  (\textbf{$C_3$}: Brain and tumor boundary)  $->$  \textbf{$C_4$}: Tumor Core)}}}
    \label{fig:brats_trail_sup}
\end{figure*}

\end{document}